\documentclass[12pt]{spieman}  
\usepackage{amsmath,amsfonts,amssymb}
\usepackage{graphicx}
\usepackage{setspace}
\usepackage{tocloft}
\usepackage{lineno}
\usepackage{color}
\usepackage{array}
\usepackage{textcomp}
\usepackage{enumitem}
\usepackage{multirow}
\usepackage{booktabs}
\usepackage{makecell}
\usepackage{algorithm}
\usepackage{algpseudocode}
\usepackage{soul}
\newcolumntype{P}[1]{>{\centering\arraybackslash}p{#1}}
\newcolumntype{M}[1]{>{\centering\arraybackslash}m{#1}}
\newcommand{\sd}[1]{{\scriptsize $\pm$ #1}}

\title{GUIDE: Guided Unbiased Image Diffusion for Endoscopy}

\author[a,*]{Siddhartha Kapuria}
\author[b]{Naruhiko Ikoma}
\author[c]{Joga Ivatury}
\author[d]{Sandeep Chinchali}
\author[a]{Farshid Alambeigi}
\affil[a]{Walker Department of Mechanical Engineering, The University of Texas at Austin, Austin, 78712, TX, USA}
\affil[b]{Department of Surgical Oncology, Division of Surgery, The University of Texas MD Anderson Cancer Center, Houston, 77030, TX, USA}
\affil[c]{Department of Surgery \& Perioperative Care, Dell Medical School, The University of
Texas at Austin, Austin, 78712, TX, USA}
\affil[d]{Chandra Family Department of Electrical and Computer Engineering, The University of Texas at Austin, Austin, 78712, TX, USA}

\cftpagenumbersoff{figure}
\cftpagenumbersoff{table} 
\begin{document} 
\maketitle

\begin{abstract}

\textbf{Purpose:} Deep learning has shown strong promise in gastrointestinal (GI) imaging, but progress is often limited by scarce and imbalanced datasets, leading to poor generalization and potential diagnostic errors. To address these challenges, we propose GUIDE: Guided Unbiased Image Diffusion for Endoscopy, a targeted data augmentation framework that combines uncertainty-based sample selection with class- and image-conditioned diffusion-based generation to improve downstream classification using only sparse class labels.

\textbf{Approach:} GUIDE is built on a class-conditioned latent diffusion model with class conditioning via cross-attention and a guided sampling strategy that injects a reference image at a tunable denoising extent to control the fidelity–diversity trade-off. An uncertainty-guided augmentation strategy identifies training samples where a baseline classifier is most uncertain, and generates multiple synthetic variations of these hard examples. We validate GUIDE on two datasets across complementary modalities: a public RGB colonoscopy dataset of colorectal polyps and a vision-tactile dataset of gastric cancer lesion phantoms. Three classifier architectures (Swin Transformer, Dilated ResNet, ConvNeXt) were evaluated under low-data regimes.

\textbf{Results:} Across both domains, GUIDE achieves high generative quality, preserves clinically relevant structures under limited data, and provides controllable diversity for augmentation. Uncertainty-guided augmentation yielded statistically significant improvements (p $<$ 0.05) across multiple architectures, with gains of up to 9.8\% in accuracy and 0.086 in F1-score, consistently outperforming real-only training, non-generative augmentation, and class-only guided synthetic data augmentation. 

\textbf{Conclusions:} GUIDE thus establishes a data-efficient and broadly applicable framework for generating diagnostically useful synthetic images, mitigating class imbalance and hard-example scarcity, and enhancing generalization in GI imaging.

\end{abstract}

\keywords{Guided Diffusion, Gastrointestinal Cancer, Synthetic Data, Vision-based Tactile Sensing}

{\noindent \footnotesize\textbf{*}Siddhartha Kapuria,  \linkable{skapuria@utexas.edu} }

\begin{spacing}{1}   

\section{Introduction}

Cancers of the gastrointestinal (GI) tract, notably colorectal cancer (CRC) and gastric cancer (GC), present a rising global health challenge. In 2020, CRC was the third most common malignancy worldwide, with roughly 1.9 million new cases and 935,000 deaths, and is projected to rise to 3.2 million cases by 2040 as populations age \cite{Sung2021GlobalCS}. GC, on the other hand, is the fifth most common cancer worldwide and the fourth leading cause of cancer-related deaths, with up to 62\% of cases diagnosed at advanced stages \cite{seer2021Cancer}. Patient survivability for these cancers has been shown to be intrinsically linked to the tumor stage at the time of diagnosis, making early detection through endoscopic procedures (colonoscopy for CRC and upper endoscopy for GC) critically important \cite{Jemal2002CancerS2}. 

Clinical assessment through such procedures relies on visualizing tumor morphology. Early‐stage CRC is characterized by features of polyps (shape, size, surface texture), while GC tumors are classified by Borrmann types I–IV based on growth patterns and ulceration, and these morphological characteristics guide diagnosis and treatment planning \cite{Axon2005UpdateOT, Kudo1994ColorectalTA, Gore2013StomachMT, Lockhart2009EpidemiologyOG}. However, the reliability of colonoscopy is constrained by two factors. First, visual diagnosis based on camera feed alone is highly subjective due to the significant morphological variation in lesions. Second, the procedure itself is limited by technical challenges, including difficult maneuverability, occlusions, and poor lighting. Consequently, diagnostic accuracy is heavily dependent on clinician expertise \cite{Ko2010ComplicationsOC}. For CRC, miss rates are approximately 26\% for adenomas, 27\% for serrated lesions, and 34\% for flat adenomas \cite{Ko2010ComplicationsOC, Zhao2019MagnitudeRF}. These missed lesions represent a critical failure in the screening process, potentially allowing pre-cancerous tissue to progress to malignancy.

To address these limitations and enhance the reliability of colonoscopy, Computer-Aided Diagnosis (CADx) systems, particularly those built on deep learning (DL) models, have risen in prominence in both endoscopic and histopathological imaging \cite{Gao2024LooseLL, Ribeiro2016ExploringDL, Li2018DeepLB,Poon2020AIdoscopistAR, Hirasawa2018ApplicationOA, Huang2021AccurateDA}. These artificial intelligence (AI) methods have shown considerable promise in automatically detecting and classifying polyps, acting as a valuable ``second observer'' to assist clinicians and reduce inter-operator variability. However, the performance and generalizability of such models are fundamentally dependent on access to large, diverse, and meticulously annotated datasets for training. This data dependency introduces a significant bottleneck in the medical imaging domain, where the acquisition of such large, balanced, and labeled datasets is notoriously difficult \cite{Ribeiro2016ExploringDL}. The process is hampered by patient privacy regulations, the high cost of data storage and curation, and the immense time and specialized expertise required for accurate annotation by clinical experts \cite{Krause2021DeepLD}. Models trained on limited or homogeneous datasets are highly susceptible to overfitting and inheriting biases present in the data \cite{Fotouhi2019ACD,Kazerouni2022DiffusionMI}. This often results in poor generalization when the model is deployed in a new clinical setting or used on patient populations different from the training data. To overcome this data scarcity problem, researchers have increasingly turned to generative models to create synthetic medical images for data augmentation. Methodologies like Generative Adversarial Networks (GANs) \cite{Goodfellow2014GenerativeAN} and, more recently, Denoising Diffusion Probabilistic Models (DDPMs) \cite{Ho2020DenoisingDP} have been successfully used to generate realistic synthetic colonoscopy images, with the goal of enhancing dataset diversity and improving the robustness of downstream classification and detection models \cite{Du2023ArSDMCI,Jeong2024UncertaintyAwareDA,Kapuria2024EnhancingCC}.

While powerful, the practical utility of many existing generative approaches is constrained by their reliance on strong conditioning information to guide the synthesis process. Several state-of-the-art methods require detailed, pixel-level segmentation masks or rich, descriptive text labels to ensure the clinical relevance and anatomical correctness of the generated images, yet such conditioning information may not always be available \cite{Fagereng2022PolypConnectII,Yoon2022ColonoscopicIS,Sharma2025DiverseIG}. Such detailed annotations are a luxury, often unavailable in the large-scale datasets used in real-world clinical research, which are frequently annotated with only high-level class labels. Medical imaging datasets are also characterized by uneven data distributions and a lack of diversity, which limits the ability of unguided or weakly-guided generative models to produce clinically useful synthetic data, particularly for rare or underrepresented polyp subtypes \cite{Fotouhi2019ACD,Krause2021DeepLD}. A critical gap, therefore, exists for a framework that can perform targeted, controllable image synthesis using only the sparse annotations commonly available in clinical practice, thereby bridging the gap between the potential of generative models and the practical realities of medical data.

To collectively address these issues, in this work, we introduce Guided Unbiased Image Diffusion for Endoscopy (GUIDE). GUIDE is a targeted data augmentation framework that first identifies training samples where a baseline model exhibits the highest uncertainty. Subsequently, a class-conditioned diffusion-based generative model creates novel variations of these challenging samples, using the original image as a structural foundation. This synthesis process is governed by a single parameter, the denoising extent, which controls the trade-off between fidelity to the input image and generative diversity. To demonstrate its versatility and effectiveness, we validate GUIDE on two distinct datasets across different modalities: public RGB endoscopic images and a unique vision-tactile dataset. For both datasets, our evaluation includes a comprehensive assessment of generative quality, the fidelity-diversity tradeoff, and performance on a downstream classification task. We show that targeted data augmentation with GUIDE significantly improves classifier performance across both modalities, highlighting its broad applicability. Importantly, our contribution lies in synthesizing established generative and active learning methods into a practical annotation-efficient data augmentation framework, enabling fine-grained image control from sparse clinical labels to improve downstream model performance.

\section{Related Works}
\label{sec:works}
\subsection{CADx in GI Imaging} 
DL–based CADx has become an important tool for automated GI imaging analysis, aiming to reduce operator dependency in lesion detection and classification. In colonoscopy, deep convolutional neural networks (CNNs) and more recently, transformers have achieved high accuracy in polyp detection and characterization \cite{Gao2024LooseLL,Ribeiro2016ExploringDL,Li2018DeepLB,Poon2020AIdoscopistAR}. For instance, a deep-learning system by Poon \textit{et al.} localized 96.9\% of polyps in colonoscopy videos, suggesting that AI assistance could help identify approximately one additional patient with polyps for every 20--33 procedures \cite{Poon2020AIdoscopistAR}. In gastric endoscopy, Hirasawa \textit{et al.} \cite{Hirasawa2018ApplicationOA} trained a CNN on over 13,000 images to automatically detect early gastric cancers, reporting 92.2\% sensitivity and near-perfect detection of lesions $\geq$6 mm. These successes demonstrate the potential of AI to augment clinician performance and mitigate variability. CADx systems have also been explored beyond endoscopy; for example, deep learning models have shown promise in histopathological analysis of colorectal cancer biopsies and surgical specimens \cite{Litjens2017Survey}. Furthermore, in our previous works, we have used a unique vision-based tactile sensing modality to demonstrate the effectiveness of both CNN and transformer-based architectures for GI cancer classification (CRC and GC) \cite{Kapuria2023TowardsRC, Kapuria2024CRF, Kapuria2024RobotEnabledGI, Kapuria2025SyntheticDE}. Across modalities, however, these models all share an underlying limitation: their accuracy and robustness are intrinsically linked to the scale and quality of training datasets, which in medicine are often limited, imbalanced, or inconsistently annotated.

\subsection{Generative Models for Medical Image Synthesis} 
Generative models offer a promising solution to expand and diversify limited medical datasets. GANs \cite{Goodfellow2014GenerativeAN} train two networks in competition (generator vs. discriminator) to produce realistic synthetic data, while DDPMs \cite{Ho2020DenoisingDP} formulate image generation as an iterative denoising process that progressively refines random noise into a coherent image. Both approaches have been leveraged to synthesize high-fidelity medical images in recent years. In GI endoscopy, Yoon \textit{et al.} \cite{Yoon2022ColonoscopicIS} employed a GAN to generate colonoscopy images containing sessile serrated lesions (a polyp subtype often missed due to flat morphology). Expert evaluation confirmed the realism of these GAN-generated images, and when used to augment training, the resulting polyp detection system showed a 17.5\% sensitivity improvement for sessile serrated lesions. Diffusion models have been similarly adopted for endoscopic image generation. For example, Du \textit{et al.} \cite{Du2023ArSDMCI} proposed an Adaptive Refinement Semantic Diffusion (ArSDM) approach to synthesize colonoscopic images from segmentation maps, enabling targeted augmentation for polyp detection and segmentation tasks. Similarly, mask-conditioned diffusion models such as Polyp-DDPM generate realistic polyp-containing images guided by a given binary mask, and have demonstrated enhanced performance in downstream segmentation \cite{Jeong2024UncertaintyAwareDA}. These works illustrate the growing trend of using GANs and diffusion models to produce additional samples that reflect the visual complexity of real endoscopic data, from color variations to diverse polyp shapes. In our previous works, we have extended the application of diffusion models for synthetic data generation of vision-based surface tactile images of both CRC and GC tumors \cite{Kapuria2024EnhancingCC, Kapuria2025AddressingBI, Kapuria2025SyntheticDE}.

\subsection{Synthetic Data Augmentation} 
The integration of synthetic images into training pipelines (i.e. data augmentation via generation) has yielded encouraging results in GI cancer applications. Augmenting limited datasets with realistic synthetic data can improve model generalization by exposing networks to greater variability. For instance, Fagereng \textit{et al.} \cite{Fagereng2022PolypConnectII} introduced PolypConnect, an image inpainting pipeline that converts normal colonoscopy frames into ones with polyps. By training a polyp segmentation model on a mix of real and PolypConnect-generated images, they achieved a 5.1\% boost in mean IoU compared to using real data alone. In another study, Golhar \textit{et al.} \cite{Golhar2025GAN} leveraged GAN inversion to expand a polyp image dataset: latent representations of pairs of real images (sharing the same polyp class) were interpolated to synthesize new samples, thereby increasing the diversity of lesion appearances. This approach improved deep CNN classification performance by 2.7\% (F1-score) and 4.9\% (sensitivity) and even enhanced robustness on an unseen hospital dataset. Such results confirm that targeted synthetic augmentation can address class imbalances (e.g. underrepresented polyp types) and reduce overfitting, ultimately strengthening CADx models.

Despite this progress, current generative augmentation methods often face practical limitations due to their reliance on dense conditioning information. Many state-of-the-art pipelines require additional annotations or inputs to guide image synthesis towards clinically valid outputs. For example, mask-to-image methods like PolypConnect \cite{Fagereng2022PolypConnectII} and diffusion models (ArSDM \cite{Du2023ArSDMCI}, Polyp-DDPM \cite{Jeong2024UncertaintyAwareDA}) assume the availability of precise polyp segmentation masks to indicate where and how a polyp should appear in the scene. Although these methods can be effective, they rely on detailed pixel-level masks. Producing such annotations for large numbers of endoscopic frames is highly time-consuming and requires expert knowledge. As an alternative, some approaches explore text or attribute conditioning: Sharma \textit{et al.} \cite{Sharma2025DiverseIG} developed a text-driven diffusion model (PathoPolyp-Diff) that can generate polyp images with specified pathology and imaging modality features. However, as the authors note, such methods rely on having detailed class labels or descriptions for each image, which are not often available in routine colonoscopy datasets. In practice, many medical datasets come with only sparse labels (e.g. diagnosis or polyp type), without segmentation maps or rich textual descriptions. Despite this, weakly guided generation relying on such sparse annotations has been underexplored. 


\section{Method}
This section details the methodology behind GUIDE, our approach for targeted synthetic data generation. Our method consists of two primary components, combining established generative and active learning techniques into a novel, practical, and clinically applicable pipeline. We first describe the generative model used in GUIDE, a class-conditioned latent diffusion model that enables fine-grained control over image synthesis. We detail its architecture, including the latent diffusion process, a sequence-aware conditioning module, and a guided sampling strategy for image-to-image translation. We then present our Uncertainty-Guided Synthetic Data Augmentation (SDA) Strategy, which leverages this model. This strategy intelligently selects reference images for synthesis based on the uncertainty of a pre-trained classifier, ensuring that the generative process is focused on the most challenging and informative data samples.

\subsection{The GUIDE Diffusion Model}

\begin{figure}[t!]
    \centering
    \includegraphics[width=1.0\linewidth]{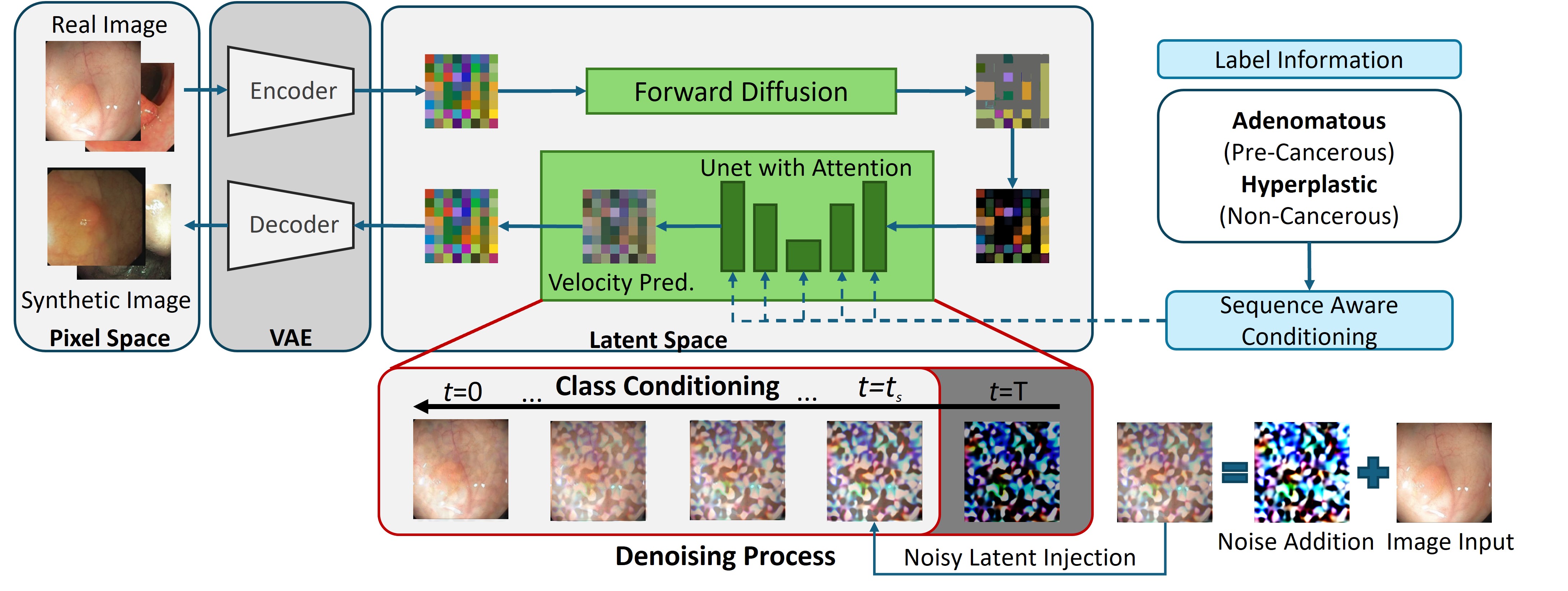}
    \caption{Overview of the GUIDE diffusion model for targeted image synthesis. The denoising process is conditioned on class labels to guide the generation. Instead of starting from \( t = 0 \), a noisy input image is injected at an intermediate timestep \( t_s \), from which the diffusion process begins, enabling controlled synthesis in the latent space.}
    \label{fig:model}
\end{figure}

The underlying generative model in GUIDE is a class-conditioned latent diffusion model (CC-LDM), built upon the foundational DDPM framework \cite{Ho2020DenoisingDP}, that operates efficiently in the latent space of a pre-trained VAE \cite{Kingma2013AutoEncodingVB}. As shown in Fig.~\ref{fig:model}, This model integrates three components to achieve controlled synthesis: (1) a latent diffusion process that serves as the generative core, (2) a class-conditioning mechanism via cross-attention that ensures tumor level consistency, and (3) a guided sampling strategy that enables targeted generation based on selected input images. These components are detailed in the following subsections.

\subsubsection{Latent Diffusion Process}
The CC-LDM operates in the latent space of a pre-trained VAE to reduce computational cost. Given an image $x_0 \in \mathbb{R}^{H \times W \times C}$, we first obtain its lower-dimensional latent representation $z_0 = \mathcal{E}(x_0)$, where $z_0 \in \mathbb{R}^{h \times w \times d}$ and $\mathcal{E}$ is the VAE encoder. The diffusion process is applied exclusively to this latent variable. The forward diffusion process gradually adds Gaussian noise  $\epsilon_t \sim \mathcal{N}(0, I)$ to the latent representation in a Markovian manner over $T$ timesteps:
\begin{align}
    q(z_t | z_{t-1}) = \mathcal{N}(z_t; \sqrt{\alpha_t} z_{t-1}, (1 - \alpha_t) I)
\end{align}
where $t \in \{1, \dots, T\}$ is the diffusion timestep and $\alpha_t \in (0,1]$ is a pre-defined noise scheduling coefficient. This process allows for sampling $z_t$ at any arbitrary timestep $t$ directly from the clean latent $z_0$:
\begin{align}
    q(z_t | z_0) = \mathcal{N}(z_t; \sqrt{\bar{\alpha}_t} z_0, (1 - \bar{\alpha}_t) I), \quad \text{where,} \;\; \bar{\alpha}_t = \prod_{i=1}^{t} \alpha_i.
\end{align}

The reverse process is parameterized by a learnable model $p_{\theta}$ that seeks to recover $z_0$ by iteratively denoising $z_T \sim \mathcal{N}(0, I)$:
\begin{align}
    p_{\theta}(z_{t-1} | z_t) = 
    \mathcal{N}\!\big(z_{t-1}; \mu_{\theta}(z_t, t, y), \Sigma_{t} I\big),
\end{align}
where $\mu_{\theta}$ is obtained from the model’s prediction, $\Sigma_{t}$ is a fixed variance term determined by the forward process, and $y$ denotes the conditioning information (the class label in our case). The specific conditioning mechanism is described in Sec.~\ref{sec:cc}.  

A common formulation is to train the network to predict the Gaussian noise $\epsilon_t$ that was added in the forward process. In this case, the mean $\mu_{\theta}$ is computed from the predicted noise $\hat{\epsilon}_t$ together with the known schedule, while the covariance $\Sigma_{t}$ is set to the analytic posterior variance. The training objective minimizes the discrepancy between $\hat{\epsilon}_t$ and $\epsilon_t$.

To improve stability and training efficiency, this noise prediction formulation can be further reparameterized as velocity prediction, as shown by Salimans and Ho \cite{salimans2022progressive}, where $v_t = (\sqrt{\bar{\alpha}_t} z_t - z_0) / \sqrt{1 - \bar{\alpha}_t}$.  
This velocity is predicted by a time-conditional UNet architecture, $v_{\theta}(z_t, t, y)$, which incorporates cross-attention, a standard choice for diffusion-based models. The model is trained by minimizing the mean squared error between the predicted and target velocities:
\begin{align}
    \mathcal{L}_{\text{LDM}} = \mathbb{E}_{z_0, y, \epsilon_t, t} \Big[ \| v_{\theta}(z_t, t, y) - v_t \|^2 \Big]
\end{align}

While this reverse process is defined in discrete timesteps, its continuous-time limit can be described by a specific ordinary differential equation (ODE) known as the probability flow ODE \cite{song2020scorebased}. This deterministic formulation is pivotal, as it recasts the generative process as an initial value problem. Solving this ODE from a latent variable of pure noise ($z_T$) at time $T$ to time $0$ yields a clean data sample ($z_0$). This perspective enables the use of advanced numerical ODE solvers to dramatically accelerate sampling. We use this formulation in our choice of sampler for the iterative denoising process (see Sec. \ref{sec:gsa}).

\subsubsection{Class Conditioning via Cross-Attention}
\label{sec:cc}
To condition the diffusion process on the polyp class, we adopt a cross-attention mechanism inspired by the text-conditioning used in large-scale models such as Stable Diffusion \cite{Rombach2021HighResolutionIS}. Instead of using a static class embedding (i.e., a single vector per class), our approach transforms the class label into a learnable embedding  matrix 
$\mathbf{E} \in \mathbb{R}^{(K+1) \times d}$. The first $K$ indices (from $0$ to $K-1$) correspond to the actual class labels. The final index, $K$, is reserved for a dedicated null token ($\emptyset$) used to enable classifier-free guidance. 

Given a label $y$ (or the null token), the corresponding vector $\mathbf{E}[y]$ is replicated $L$ times to form a conditioning matrix. This matrix is then combined with a second learnable positional embedding, $\mathbf{P} \in \mathbb{R}^{L \times d}$, via element-wise addition. This produces the final conditioning representation $C(y) \in \mathbb{R}^{L \times d}$, which encodes both class (or null) identity and positional context.

This conditioning representation $C(y)$ is integrated into the UNet architecture via cross-attention layers at multiple spatial resolutions. The attention operation, which allows the model's image features to selectively draw information from the conditioning representation $C(y)$, is formulated as scaled dot-product attention:
\begin{align}
\text{Attention}(Q, K, V) = \text{softmax}\left(\frac{QK^T}{\sqrt{d_k}}\right)V
\end{align}
In this formulation, the Query ($Q$) is a linear projection of the UNet's intermediate spatial features. The Key ($K$) and Value ($V$) are linear projections of the conditioning representation $C(y)$, and $d_k$ is the dimension of the key vectors. This allows the model to apply class-specific guidance in a spatially-aware manner, providing more flexible and powerful conditioning than a single global vector. When $\emptyset$ is provided as input, the model performs unconditional image generation.

\subsubsection{Guided Sampling Strategy:}
\label{sec:gsa}
Our guided sampling strategy enables controlled image-to-image translation, leveraging an input image as a structural prior. Given an input image $x_{\text{in}}$, we first obtain its latent representation $z_{\text{in}}$ using the VAE encoder. This latent is then perturbed by a partial forward diffusion process up to a specific timestep $t_s$:
\begin{align}
z_{t_s} = \sqrt{\bar{\alpha}{t_s}} z_{\text{in}} + \sqrt{1 - \bar{\alpha}_{t_s}} \epsilon, \quad \text{where,}\;\;  \epsilon \sim \mathcal{N}(0, I).
\end{align}
The injection step $t_s$ is a crucial hyperparameter that balances fidelity and diversity in the generation process. To control this injection intuitively, we introduce a tunable parameter termed the \textit{denoising extent}, denoted by $d_{\text{ext}} \in [0,1]$, which linearly maps to the inference timesteps of the scheduler as:
\[
t_s = \lfloor (1-d_{\text{ext}})(T-1)\rfloor.
\]
A small $d_{\text{ext}}$ ($d_{\text{ext}} \approx 0$) corresponds to starting from nearly clean latents and thus preserves more structure from $x_{\text{in}}$, while a large $d_{\text{ext}}$ ($d_{\text{ext}} \approx 1$) corresponds to starting from highly noisy latents, allowing for more diverse synthesis guided primarily by the class label. Setting $d_{\text{ext}}=1$ is equivalent to unconditional sampling from pure noise. We adopt this simple linear mapping for clarity and reproducibility, since our experiments fix both the scheduler and the number of inference steps.

Starting from the noisy latent $z_{t_s}$, the reverse diffusion process is guided using Classifier-Free Guidance (CFG) to steer the generation toward the target class. At each step, the UNet computes two velocity predictions: a class-conditioned prediction $v_{\theta}^{(c)} = v_{\theta}(z_t, t, y)$ and an unconditioned prediction $v_{\theta}^{(u)} = v_{\theta}(z_t, t, \emptyset)$, where $\emptyset$ is a null token. The final velocity is a weighted combination of these two:
\begin{align}
v_{\theta} = v_{\theta}^{(u)} + \gamma (v_{\theta}^{(c)} - v_{\theta}^{(u)})
\end{align}
where $\gamma > 1$ is the guidance scale. For the iterative denoising, we employ the UniPC (Unified Predictor-Corrector) sampler \cite{zhao2023unipc}. UniPC is an efficient, high-order solver for the probability flow ODE \cite{song2020scorebased} that governs the reverse diffusion process. While standard stochastic samplers like DDPM typically require $T=1000$ steps to produce high-fidelity medical images, UniPC utilizes a predictor-corrector method to ensure stable and accurate sampling, allowing for the generation of high-fidelity images in significantly fewer function evaluations ($T=100$ in our implementation). We chose this sampler to significantly accelerate our guided synthesis pipeline while maintaining high sample quality. The denoised latent $z_0$ is then mapped back to the pixel space using the VAE decoder to produce the final synthetic image.

\begin{algorithm}[t!]
\caption{Uncertainty-Guided Synthetic Data Augmentation}\label{alg:augmentation}
\begin{algorithmic}[1]
\State \textbf{Input:} Original training set $\mathcal{D}_{\text{train}}$, pre-trained CC-LDM $\mathcal{G}$, number of hard samples $N$, synthetic images per sample $k$, denoising extent $d_{\text{ext}}$.
\State \textbf{Output:} Augmented training set $\mathcal{D}_{\text{aug}}$.
\Statex
\Statex \textit{Step 1: Train baseline classifier}
\State $M_{\text{base}} \gets \text{TrainClassifier}(\mathcal{D}_{\text{train}})$
\Statex
\Statex \textit{Step 2: Identify high-uncertainty samples}
\State $\mathcal{U} \gets \{\}$ \Comment{Initialize set of (uncertainty, image) pairs}
\For{each $(x, y) \in \mathcal{D}_{\text{train}}$}
    \State $p \gets M_{\text{base}}(x)$ \Comment{Get softmax probabilities}
    \State $u \gets \text{CalculateUncertainty}(p)$ \Comment{e.g., Entropy or 1-max($p$)}
    \State $\mathcal{U} \gets \mathcal{U} \cup \{(u, x, y)\}$
\EndFor
\State $\mathcal{D}_{\text{hard}} \gets \text{SelectTopN}(\mathcal{U}, N)$ \Comment{Select $N$ samples with highest uncertainty}
\Statex
\Statex \textit{Step 3: Generate synthetic data with controlled diversity}
\State $\mathcal{D}_{\text{synth}} \gets \{\}$
\For{each $(u, x_{\text{in}}, y_{\text{in}}) \in \mathcal{D}_{\text{hard}}$}
    \For{$i = 1$ \textbf{to} $k$} \Comment{Generate $k$ variations for each hard sample}
        \State $x_{\text{synth}} \gets \mathcal{G}(x_{\text{in}}, y_{\text{in}}, d_{\text{ext}})$
        \State $\mathcal{D}_{\text{synth}} \gets \mathcal{D}_{\text{synth}} \cup \{(x_{\text{synth}}, y_{\text{in}})\}$
    \EndFor
\EndFor
\Statex
\Statex \textit{Step 4: Create augmented dataset}
\State $\mathcal{D}_{\text{aug}} \gets \mathcal{D}_{\text{train}} \cup \mathcal{D}_{\text{synth}}$
\State \textbf{return} $\mathcal{D}_{\text{aug}}$
\end{algorithmic}
\end{algorithm}

\subsection{Uncertainty-Guided Synthetic Data Augmentation}
\label{sec:uga}
While the CC-LDM can generate a synthetic image from any reference, a targeted data SDA strategy is needed to maximize its impact on downstream tasks. Not all synthetic samples provide equal value; augmenting the dataset with variations of samples the model already classifies correctly is inefficient. Instead, our approach adopts a strategy inspired by active learning, specifically uncertainty sampling \cite{settles2009active}. We focus on generating new data in regions of the feature space where a baseline model is most uncertain, effectively strengthening the model where it is weakest by populating sparsely sampled areas near its decision boundary.

Algorithm 1 details the full procedure for the uncertainty-guided SDA strategy used in GUIDE. The process takes the original training set, the trained CC-LDM, the number of hard samples to select ($N$), the number of synthetic variations to create per sample ($k$), and the desired denoising extent ($d_{\text{ext}}$) as input. The procedure begins by training a baseline classification model on the original data (line 3). This model is then used to identify the $N$ training samples with the highest prediction uncertainty (lines 4-10). Considering $p$ to be the softmax probability vector, we explore two uncertainty scores. The first is based on Low Confidence (LC), calculated as $1 - \max(p)$. The second is based on High Entropy (HE), calculated using the Shannon entropy formula \cite{shannon1948mathematical}:
\begin{align}
H(p) = - \sum_{i=1}^{K} p_i \log_2(p_i)
\end{align}
Notably, while we adopt LC and HE in this work, the selection criterion can be readily replaced with any other uncertainty or informativeness metric, making the strategy broadly applicable across different tasks and domains.
In the third step, the algorithm iterates through each of these $N$ hard samples, and for each one, the CC-LDM is called $k$ times to generate multiple synthetic variations. The generation is controlled by the denoising extent parameter $d_{\text{ext}}$, which determines how much structure from the input sample is preserved versus how much diversity is introduced (see Sec.~\ref{sec:gsa}). Smaller $d_{\text{ext}}$ values retain high fidelity to the original sample, while larger values promote more diverse outputs. Finally, the resulting set of $N \times k$ synthetic images is combined with the original training set to form the final augmented dataset (line 18), which is then used to train a new, more robust classifier.

\section{Datasets}
To thoroughly evaluate the performance and generalizability of our method, we utilize two distinct datasets that span different modalities and data sources. The first is a large, publicly available collection of RGB colonoscopy images of adenomatous and hyperplastic CRC polyps. The second is a unique, vision-based tactile imaging dataset of GC data collected in-house, which allows us to test the versatility of our approach on more complex but limited vision-tactile data. Details about each dataset are provided in this section. 

\subsection{Public RGB Colonoscopy Dataset}
Our primary dataset is a large-scale collection of colonoscopy frames of adenomatous and hyperplastic CRC polyps, curated by Li et al \cite{Li2021ColonoscopyPD}. This dataset aggregates several publicly available sources, including  MICCAI 2017 (GIANA Challenge) \cite{bernal2017comparative}, CVC-ColonDB \cite{bernal2012towards}, GLRC \cite{mesejo2016computer}, and the KUMC dataset \cite{Li2021ColonoscopyPD}, each containing colonoscopy video sequences with varying levels of annotation detail. MICCAI 2017 and CVC-ColonDB are labeled with segmentation masks but lack classification information. GLRC provides class labels but no bounding box annotation, while KUMC is fully annotated with both class labels and bounding boxes. Li et al. \cite{Li2021ColonoscopyPD} provided labels for all sequences where missing, and also ensured balanced representation accounting for camera movement and video length. 
Frames with severe motion blur or illumination changes were removed, while mildly noisy frames were kept to enhance model robustness.  
The final release consists of 155 video sequences and 37899 images split \emph{by sequence} to prevent leakage with approximately 75\%/10\%/15\% train/val/test per source and per class (yielding 116/17/22 sequences and 28,773/4,254/4,872 frames, respectively). We follow the authors’ video-level split protocol in all experiments. The train set is 50/45 adenomatous vs. hyperplastic, with $\sim5\%$ unclassified due to non-distinctive nature, while the validation and test sets have equal class representation. The unclassified images were discarded during all experiments in this work. The dataset is publicly available via Harvard Dataverse (DOI: \texttt{10.7910/DVN/FCBUOR}). For training the CC-LDM, we used only the images and their corresponding label.

\begin{figure}
    \centering
    \includegraphics[width=1.0\linewidth]{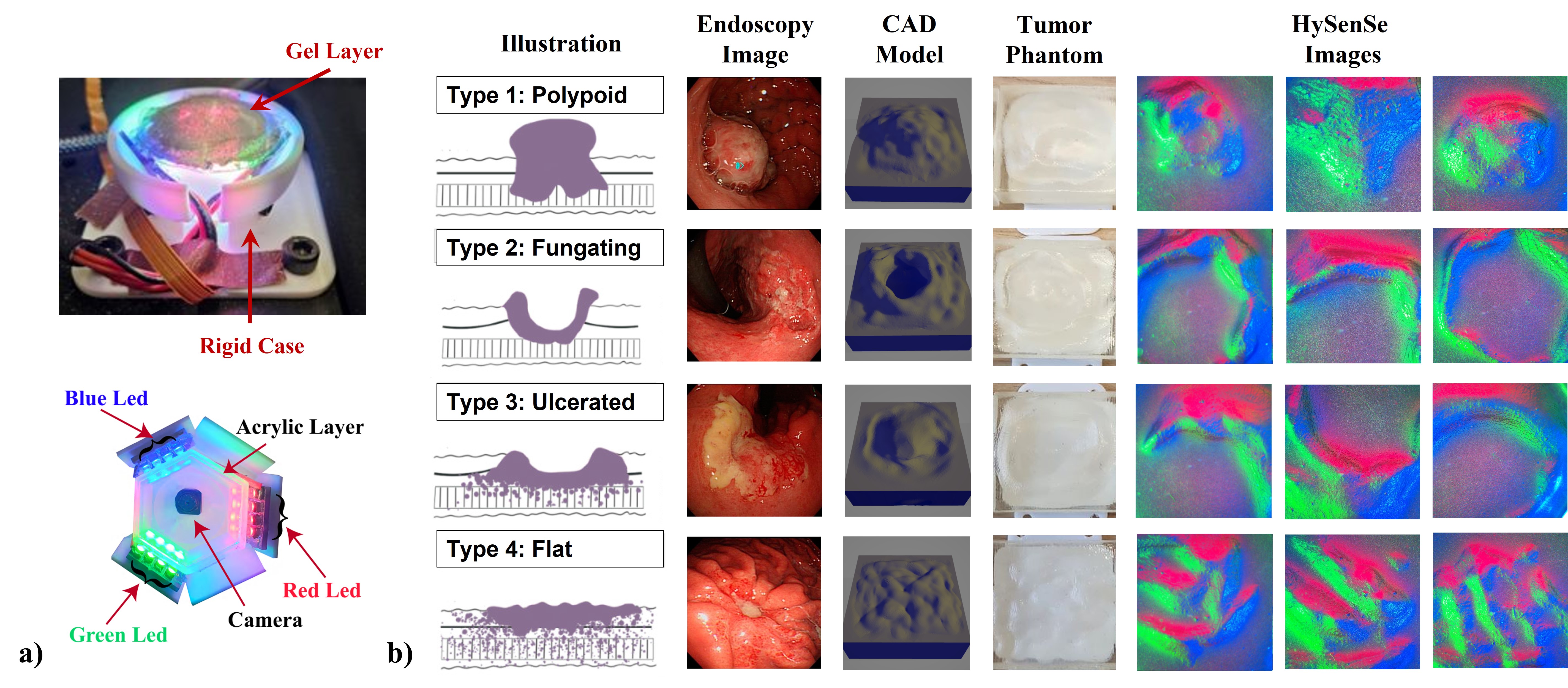}
    \caption{(a) Vision-based Surface Tactile Sensor \textit{HySenSe} and its components (b) Representative \textit{HySenSe} images of the phantom GC tumor dataset, along with their real colonoscopy images, CAD models, and 3D printed phantoms.}
    \label{fig:pgi_dataset}
\end{figure}

\subsection{Vision-Based Surface Tactile GC Phantom Dataset}
To test the adaptability of our framework to different modalities, we also use a unique vision-based surface tactile dataset of advanced GC phantoms from our previous works \cite{Kapuria2024RobotEnabledGI, Kapuria2025AddressingBI}. This dataset was collected using our novel vision-based tactile sensor (VTS), \textit{HySenSe} (see Fig. \ref{fig:pgi_dataset}a), which is designed to acquire high-fidelity textural images of tissue surfaces \cite{Kara2022HySenSeAH, Kara2023ASH}. As shown in Fig. \ref{fig:pgi_dataset}a, the sensor consists of a deformable silicone membrane, an internal camera, and an array of RGB LEDs for illumination. Its operating principle is to visually capture the deformation of the membrane as it makes contact with a surface. The internal LEDs create unique lighting patterns based on the deformation, allowing the resulting RGB image to implicitly encode physical properties like tissue stiffness and surface texture. This effectively converts rich tactile information into a single textural image that can be analyzed by standard computer vision algorithms. Notably, an inflatable version of this sensor is also under development \cite{kim2023soft}, aimed at enabling the same tactile analysis directly \textit{in vivo} within the colon.

The dataset was created using this sensor on custom-made silicone phantoms designed to mimic the biomechanical properties of various tumor types, according to the Borrmann classification of gastric cancer \cite{Gore2013StomachMT}.
Data was collected using a 7-DoF robotic manipulator (with the \textit{HySenSe} sensor as the end effector) to perform semi-autonomous contact and scanning of 44 unique additively manufactured phantoms, resulting in a total of 2,200 samples (550 per class). Each data sample is a high-resolution RGB textural image captured by the sensor's internal camera. Representative images for each class are provided in Fig. \ref{fig:pgi_dataset}b. This dataset is particularly challenging as the sensor's limited area only captures partial images of the larger tumor phantoms ($\sim 3\text{cm} \times 3\text{cm}$), allowing us to rigorously test the generalization capabilities of our data augmentation strategy. For the experiments in this study, the originally balanced dataset was artificially imbalanced to simulate a more realistic clinical distribution. The final class distribution used for training was 23.8\% for Type I, 39.2\% for Type II, 20.3\% for Type III, and 16.7\% for Type IV. The dataset was split at the tumor level using stratified sampling: 20\% was reserved for testing, while the remaining 80\% was divided into 75\% training and 25\% validation. For complete details on phantom fabrication and the robotic data collection procedure, we refer readers to our previous work \cite{Kapuria2024RobotEnabledGI}.

\section{Experiments}
To validate our approach, we conducted a series of experiments designed to answer three primary questions: (1) What are the data requirements for training the CC-LDM? (2) What is the quality of the synthetic images generated, and how is it affected by the denoising extent ($d_{\text{ext}}$)? (3) Does our Uncertainty-Guided Augmentation strategy improve the performance of a downstream classifier? We performed these evaluations on both our in-house vision-tactile dataset and the public endoscopy dataset.

\subsection{Evaluation Metrics}
\subsubsection{Generative Quality}
To evaluate the generative quality of our CC-LDM, we used a suite of standard metrics. For dataset-level assessment, we used Fréchet Inception Distance (FID) \cite{Heusel2017GANsTB} and Kernel Inception Distance (KID) \cite{Binkowski2018DemystifyingMG}. FID is a widely used metric that computes the distance between feature distributions of real and generated images using an Inception-V3 network \cite{szegedy_rethinking_2015}. To provide a more comprehensive and robust evaluation, we also report KID, which is a related metric that compares the distributions using the squared Maximum Mean Discrepancy (MMD). Unlike FID, KID does not assume the features follow a Gaussian distribution and has an unbiased estimator, making it more robust, especially on smaller sample sets. For both FID and KID, lower scores are better. Here, we must note that since the Inception-V3 network was pre-trained on natural images, its features may not perfectly capture the minute textural characteristics of specific medical data (both colonoscopy frames and VTS textural images). Therefore, we use FID and KID primarily for comparative purposes to evaluate relative performance rather than as an absolute measure of image quality.

For image-level assessment, we used Multi-Scale Structural Similarity (MS-SSIM)\cite{wang2003multiscale} to measure structural fidelity by comparing image properties like luminance and contrast at multiple scales (higher is better). We also evaluated image similarity using Learned Perceptual Image Patch Similarity (LPIPS) \cite{Zhang2018TheUE}, which compares deep features extracted from a neural network and aligns more closely with human perception of visual differences (lower values indicate greater similarity). In line with common practice, we used the AlexNet \cite{Krizhevsky2012ImageNetCW} backbone in this work.  

\subsubsection{Classification Performance}
To evaluate the performance of the downstream classifier (see Sec. \ref{sec:task} for details), we report Accuracy and the macro-averaged F1-Score. Accuracy provides a clear measure of the overall correctness of predictions. However, to provide a more robust assessment that accounts for potential class imbalance, we also report the F1-Score, which is the harmonic mean of Precision and Recall. For the multi-class vision-tactile dataset, the F1-Score was macro-averaged to ensure each class contributed equally to the final metric. To evaluate the reliability of the performance improvements, we also performed a two-tailed Welch's t-test comparing the augmented models against the corresponding ``Real Data Only" baselines. We report statistical significance ($p < 0.05$) denoted by an asterisk (*) in the results.

\subsection{Experimental Design}
\subsubsection{Data Requirements Analysis}
We first examined how the amount of training data influences the generative performance of CC-LDM. To this end, we trained models on progressively larger fractions of the available training set (5\%, 10\%, 20\%, 40\%, 80\%, and 100\%). For each configuration, we generated 1,000 synthetic images, keeping $d_{\text{ext}}=1$, and compared them against the held-out real test set using FID and KID. This analysis quantifies how data availability impacts the quality of the learned generative distribution, while assessing the full generative quality of the model without prior image information. 

\subsubsection{Generative Quality and Parameter Analysis}
\label{sec:param_analysis}
Secondly, we analyzed the fidelity–diversity trade-off introduced by the guided sampling strategy (see Sec. \ref{sec:gsa}). For each class, we randomly selected $N=50$ reference images \textit{per-class} from the test set and generated $M=5$ synthetic variations per reference at different values of the denoising extent $d_{\text{ext}}$. To ensure that statistical comparisons were made at the level of independent reference images, the similarity metrics for each set of $M$ generations were aggregated using the median, yielding one representative score for each original image. The similarity between generated samples and their corresponding references was quantified using MS-SSIM and LPIPS, providing complementary measures of structural fidelity and perceptual realism.

\subsubsection{Uncertainty-Guided Synthetic Data Augmentation}
\label{sec:task}
The third main experiment evaluated the impact of our augmentation strategy on downstream classification performance. We compared three representative classifier architectures, namely Swin Transformer \cite{liu2021swin}, Dilated ResNet \cite{Yu2017DilatedRN}, and ConvNeXt \cite{liu2022convnet}, selected to capture complementary design paradigms. The Swin Transformer represents a hierarchical vision transformer, which offers strong modeling of long-range dependencies but is often data-hungry and computationally heavy \cite{liu2021swin}. The Dilated ResNet is a residual CNN variant with dilated convolutions, which improves receptive field coverage and efficiency but generally achieves lower accuracy than modern backbones \cite{He2016DeepRL, Yu2017DilatedRN}. ConvNeXt is a recent convolutional architecture that incorporates design elements inspired by transformers, providing a favorable balance of accuracy and efficiency while remaining more robust than standard CNNs under limited data \cite{liu2022convnet}.  

Each classifier was trained on both datasets under four distinct configurations: (1) Real Data Only (Baseline): training exclusively on the original dataset without synthetic augmentation; (2) HE/LC Oversampling (HE/LC OS): a non-generative baseline where the top 10\% ``hard'' samples (based on HE or LC, see Sec.~\ref{sec:uga}) were oversampled from the existing real data; (3) Class-Conditioning Only: augmentation with synthetic images generated solely from class labels, corresponding to starting the generative denoising process from pure noise ($d_{\text{ext}}=1$) and producing samples that capture class-level semantics but lack instance-specific information; and (4) Uncertainty-Guided SDA (Image+Class Conditioned): augmentation guided by the top 10\% of ``hard'' training samples (HE or LC). 
For each such sample, $k=4$ synthetic variations were generated while varying the denoising extent $d_{\text{ext}}\in\{0.3,0.5,0.7\}$ to explore the effect of the fidelity–diversity trade-off on the classification task. This design enabled us to directly compare our targeted, image-conditioned augmentation against both the real-only baseline and the simpler class-only approach. It is important to note that while test set images were used as external references to probe the fidelity–diversity trade-off in the previous experiment (Sec. \ref{sec:param_analysis}), no test-anchored generations were ever introduced into model training or augmentation pipelines for the classification task. Furthermore, while we benchmark our framework against real-only training and class-only augmentation, we avoid comparisons to mask- or text-conditioned diffusion methods (e.g.,  ArSDM \cite{Du2023ArSDMCI}, Polyp-DDPM \cite{Jeong2024UncertaintyAwareDA}), since they rely on dense pixel-level annotations or descriptive metadata that are often unavailable in clinical datasets. Instead, we position GUIDE as a practical and annotation-efficient alternative for scenarios where the expert effort for such detailed annotations is prohibitive or unavailable.

\subsubsection{Component Ablation}
\label{sec:ablation}
To evaluate the contribution of each component in GUIDE, we also conducted a systematic ablation study isolating the effects of uncertainty-guided sample selection, generative augmentation, and image-guided synthesis. We considered four experimental configurations that reflect the progression from conventional training to the full guided diffusion framework. To maintain a controlled ablation setting, we fixed the uncertainty criterion to HE and the guidance strength to ($d_{\text{ext}}=0.5$) throughout. Specifically, we evaluated: (1) a baseline trained only on real data; (2) a sample-selection-only setting using HE oversampling (HE OS), representing conventional uncertainty-based resampling without any generative augmenation; (3) a generative SDA setting using only label-conditioned generation ($d_{\text{ext}}=1.0$), to assess the value of synthetic augmentation without image guidance; and (4) the full GUIDE configuration, which combines uncertainty-based sample selection, generative augmentation, and image-guided synthesis. These configurations were evaluated across all three backbones and both datasets to assess the consistency of component contributions across architectures and domains.

\subsection{Implementation Details}
This sub-section provides the key architectural choices and training parameters used in our experiments to ensure reproducibility. All experiments were performed on a system with an Intel Xeon Gold 6346 CPU (3.10GHz) and an NVIDIA RTX 6000 Ada (48GB VRAM). Models were implemented in PyTorch 2.7.0 with CUDA 12.6.

\subsubsection{Diffusion Model}
All images used for training the CC-LDM model were resized to $256 \times 256$ pixels and normalized to a $[0, 1]$ range. We used the same weights for the VAE as Stable Diffusion 1.5 \cite{Rombach2021HighResolutionIS}, sourced from the Huggingface Diffusers library \cite{von-platen-etal-2022-diffusers}. This VAE consists of four downsampling blocks, reducing the input resolution by a factor of 8 and mapping $256 \times 256$ images to a latent space of size $4 \times 32 \times 32$. For our sequence-aware conditioning, the embedding dimension $d$ was set to 256 and the sequence length $L$ was 8. The model was trained for 100 epochs with a batch size of 8 and an initial learning rate of 0.0001, using the AdamW optimizer, a OneCycleLR scheduler, and a squared cosine noise schedule. An Exponential Moving Average (EMA) of the model weights with a decay of 0.999 was maintained during training.  During inference, we used the UniPC sampler with 100 steps and an empirically set CFG scale of $\gamma=3.5$, sampling with the EMA weights to ensure high-quality output. Only the denoising extent parameter ($d_{\text{ext}}$) was changed according to the experiment. Following the recommendations of Lin et al. \cite{lin2024common}, we also modified the timestep spacing to be ``trailing''. This approach concentrates the sampling steps towards the end of the denoising process (i.e., near $t=0$), which has been shown to improve the generation of fine-grained details.

\subsubsection{Downstream Classification Task}
For classification experiments, we used the base implementations of Swin Transformer, Dilated ResNet, and ConvNeXt as provided by \texttt{torchvision} \cite{torchvision2016}, with ImageNet-1K \cite{Deng2009ImageNetAL} pretrained weights. To emphasize augmentation effects, we simulated a low-data regime through random stratified selection: for the RGB colonoscopy dataset, 2,000 training and 400 validation images; for the VTS GC Phantom dataset, 500 training and 200 validation images, with the test sets unchanged. Importantly, these splits were reused for all experiments to ensure fair comparisons and prevent leakage. The final classification head of each model was modified to match the number of classes in the respective dataset. During fine-tuning, we applied a series of standard data augmentations to enhance robustness, including random cropping, horizontal and vertical flips, random rotations up to 45 degrees, and Gaussian noise/blur. All classifiers were trained for up to 100 epochs using the Adam optimizer, a batch size of 32, a learning rate of 0.0001, and an early stopping patience of 10 epochs.

\section{Results and Discussion}
\subsection{Training Data versus Model Performance}

\begin{figure}[t]
    \centering
    \includegraphics[width=1\linewidth]{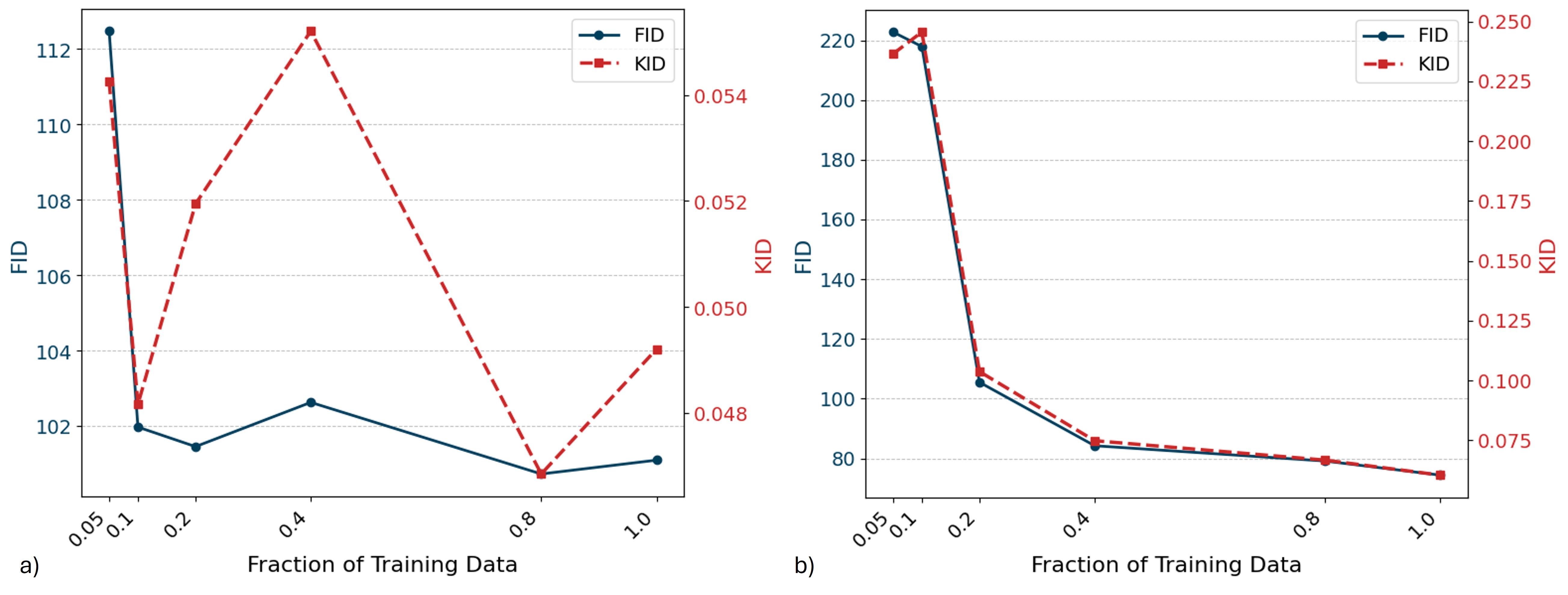}
    \caption{Dataset metrics versus amount of training data}
    \label{fig:data_req}
\end{figure}

Figure \ref{fig:data_req} shows the effect of training set size on the generative quality of the CC-LDM, measured by FID and KID. On the colonoscopy dataset (Fig. \ref{fig:data_req}a), which is substantially larger, FID decreases from $\approx$112 at 5\% of the data to $\approx$102 at 20\%, indicating a modest improvement in alignment with the real distribution as more data becomes available. Interestingly, this drop is not mirrored in KID, which remains relatively stable ($\approx$0.045–0.055) across all fractions. One possible explanation is that FID, which assumes Gaussian feature distributions, is more sensitive to small shifts in mean and covariance when the sample size is large, whereas KID (based on unbiased MMD) remains more stable in these settings. This suggests that FID may overstate improvements in large-sample regimes, while KID provides a more conservative estimate. In contrast, results on the smaller VTS GC Phantom dataset (Fig. \ref{fig:data_req}b) show a much stronger dependence on data availability. Both FID and KID exhibit higher divergence relative to the full dataset when only 5\% of the data is used (FID $>$ 200, KID $\approx$ 0.24) and decrease monotonically as more data is added, reaching FID $\approx$ 80 and KID $\approx$ 0.07 with the full dataset. This contrast can be attributed to the dataset scale: since the VTS GC Phantom dataset is severely limited, reductions in training fraction have a more pronounced effect on CC-LDM performance, and the most stable generative quality is observed when the full dataset is used. 

\subsection{Synthetic Image Generation}
\subsubsection{Qualitative Evaluation}

\begin{figure}
    \centering
    \includegraphics[width=1.0\linewidth]{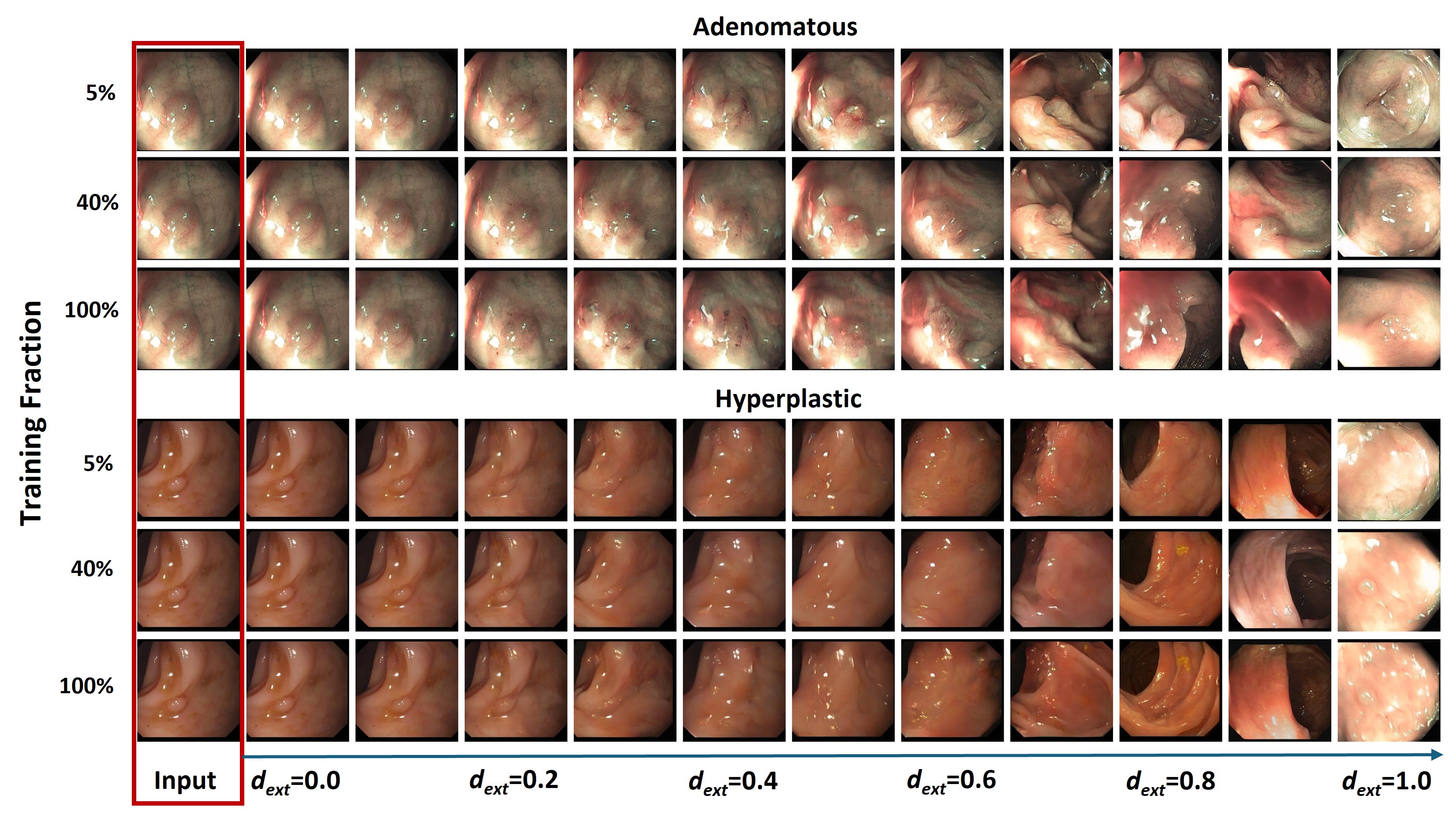}
    \caption{Representative images generated by the CC-LDM for selected reference images for the RGB colonoscopy dataset, showing the effect of $d_{\text{ext}}$ on image characteristics and quality across different training fractions.}
    \label{fig:crc-synth}
\end{figure}

\begin{figure}
    \centering
    \includegraphics[width=1.0\linewidth]{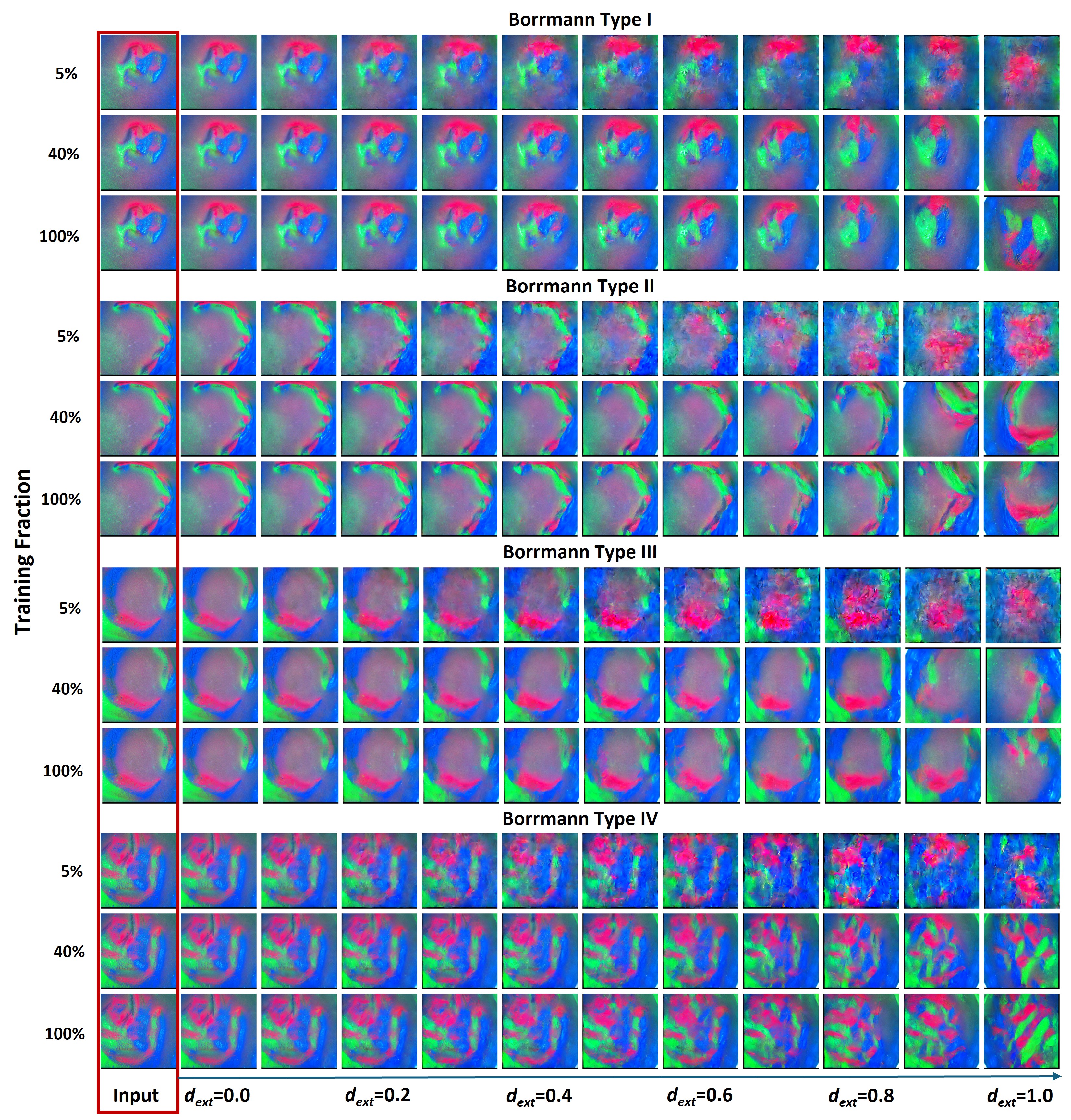}
    \caption{Representative images generated by the CC-LDM for selected reference images for the VTS GC Phantom dataset, showing the effect of $d_{\text{ext}}$ on image characteristics and quality across different training fractions.}
    \label{fig:pgi-synth}
\end{figure}

\label{sec:qualitative}
Figures~\ref{fig:crc-synth} and \ref{fig:pgi-synth} present representative synthetic images for both datasets, showing per-class synthetic images generated by varying the denoising extent $d_{\text{ext}}$. With 100\% training data, images generated at high $d_{\text{ext}}$ (closer to 1.0) appear largely randomized, reflecting class-level semantics without structural guidance from the reference. As $d_{\text{ext}}$ decreases, the synthetic samples progressively adopt the spatial and textural characteristics of the input, with tumor-level features becoming evident at intermediate values (e.g., $d_{\text{ext}}=0.5$) and fine-grained textures and lighting conditions preserved at lower values (e.g., $d_{\text{ext}}=0.3$). When the model is trained on reduced fractions of data, we observe stronger reliance on the reference image for spatial details, with salient features emerging even at higher $d_{\text{ext}}$. However, this setting also introduces more artifacts and potentially unrealistic distortions, reflecting reduced generalizability under limited training. In contrast, with sufficient training data, later denoising stages (higher $d_{\text{ext}}$) yield diverse yet faithful reconstructions of class characteristics across both datasets. For the VTS GC Phantom dataset, realistic tumor appearances emerge only at low $d_{\text{ext}}$ where the image prior dominates, while at higher $d_{\text{ext}}$ the model fails to generate plausible structures, highlighting degraded performance under data-scarce training. Nevertheless, even under limited data, low $d_{\text{ext}}$ remains effective, as the stronger influence of the reference image helps maintain realism and ensures useful synthetic variations.

\begin{figure}
    \centering
    \includegraphics[width=1.0\linewidth]{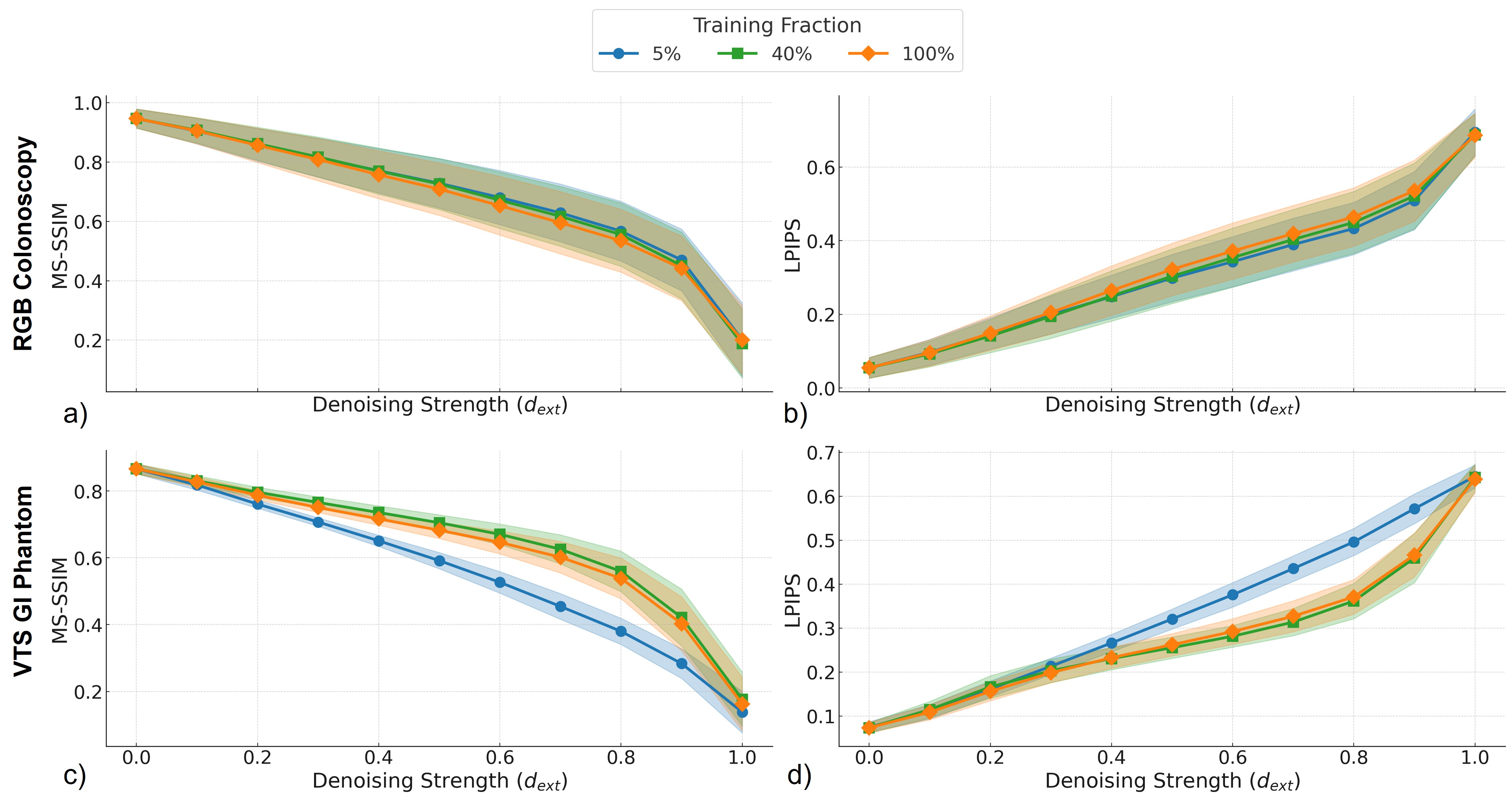}
    \caption{Fidelity–diversity trade-off across denoising extent $d_{\text{ext}}$}
    \label{fig:fidelity_diversity}
\end{figure}

\subsubsection{Quantitative Evaluation}
\label{sec:quantitative}

\begin{table}[t]
\centering
\small
\caption{MS-SSIM and LPIPS (mean $\pm$ std.) at different $d_{\text{ext}}$ and training fractions for the RGB Colonoscopy dataset.}
\label{tab:crc_sim}
\setlength{\tabcolsep}{4pt}
\begin{tabular}{cc cc cc}
\toprule
\multirow{2}{*}{\textbf{$d_{\text{ext}}$}} & \multirow{2}{*}{\shortstack{\textbf{Training} \\ \textbf{Fraction}}}
 & \multicolumn{2}{c}{\textbf{MS-SSIM $\uparrow$}} & \multicolumn{2}{c}{\textbf{LPIPS $\downarrow$}} \\
\cmidrule(lr){3-4} \cmidrule(lr){5-6}
 & & \textbf{Adenomatous} & \textbf{Hyperplastic} & \textbf{Adenomatous} & \textbf{Hyperplastic} \\
\midrule
0.1 & 5\%   & 0.915 \sd{0.041} & 0.898 \sd{0.046} & 0.093 \sd{0.037} & 0.100 \sd{0.036} \\
0.1 & 40\%  & 0.917 \sd{0.040} & 0.901 \sd{0.046} & 0.088 \sd{0.037} & 0.096 \sd{0.037} \\
0.1 & 100\% & 0.917 \sd{0.040} & 0.898 \sd{0.048} & 0.089 \sd{0.036} & 0.100 \sd{0.038} \\
\midrule
0.5 & 5\%   & 0.756 \sd{0.078} & 0.708 \sd{0.082} & 0.296 \sd{0.066} & 0.297 \sd{0.056} \\
0.5 & 40\%  & 0.743 \sd{0.083} & 0.714 \sd{0.090} & 0.311 \sd{0.078} & 0.297 \sd{0.069} \\
0.5 & 100\% & 0.750 \sd{0.077} & 0.719 \sd{0.090} & 0.293 \sd{0.073} & 0.297 \sd{0.065} \\
\midrule
1.0 & 5\%   & 0.183 \sd{0.115} & 0.181 \sd{0.114} & 0.643 \sd{0.053} & 0.648 \sd{0.049} \\
1.0 & 40\%  & 0.179 \sd{0.123} & 0.188 \sd{0.114} & 0.653 \sd{0.058} & 0.633 \sd{0.057} \\
1.0 & 100\% & 0.159 \sd{0.125} & 0.167 \sd{0.124} & 0.655 \sd{0.059} & 0.634 \sd{0.056} \\
\bottomrule
\end{tabular}
\end{table}

\begin{table}[t]
\centering
\caption{MS-SSIM and LPIPS (mean $\pm$ std.) for VTS GC Phantom dataset across different $d_{\text{ext}}$ and training fractions.}
\label{tab:pgi_sim}
\small
\setlength{\tabcolsep}{3.5pt}

\begin{tabular}{cc cc cc}
\toprule
\multirow{2}{*}{\textbf{$d_{\text{ext}}$}} & \multirow{2}{*}{\shortstack{\textbf{Training} \\ \textbf{Fraction}}}
 & \multicolumn{2}{c}{\textbf{MS-SSIM $\uparrow$}} & \multicolumn{2}{c}{\textbf{LPIPS $\downarrow$}} \\
\cmidrule(lr){3-4} \cmidrule(lr){5-6}
 &  & \textbf{Type I} & \textbf{Type II} & \textbf{Type I} & \textbf{Type II} \\
\midrule
0.1 & 5\%   & 0.822 \sd{0.011} & 0.834 \sd{0.011} & 0.123 \sd{0.014} & 0.108 \sd{0.012} \\
0.1 & 40\%  & 0.824 \sd{0.011} & 0.836 \sd{0.011} & 0.121 \sd{0.013} & 0.106 \sd{0.012} \\
0.1 & 100\% & 0.825 \sd{0.011} & 0.836 \sd{0.010} & 0.121 \sd{0.013} & 0.106 \sd{0.011} \\
\midrule
0.5 & 5\%   & 0.645 \sd{0.015} & 0.657 \sd{0.014} & 0.350 \sd{0.017} & 0.333 \sd{0.016} \\
0.5 & 40\%  & 0.650 \sd{0.015} & 0.661 \sd{0.014} & 0.345 \sd{0.016} & 0.329 \sd{0.016} \\
0.5 & 100\% & 0.652 \sd{0.015} & 0.662 \sd{0.014} & 0.345 \sd{0.016} & 0.329 \sd{0.016} \\
\midrule
1.0 & 5\%   & 0.158 \sd{0.014} & 0.167 \sd{0.014} & 0.617 \sd{0.015} & 0.603 \sd{0.013} \\
1.0 & 40\%  & 0.160 \sd{0.014} & 0.168 \sd{0.014} & 0.616 \sd{0.015} & 0.602 \sd{0.013} \\
1.0 & 100\% & 0.161 \sd{0.014} & 0.169 \sd{0.014} & 0.616 \sd{0.015} & 0.602 \sd{0.013} \\
\bottomrule
\end{tabular}

\vspace{1em} 

\begin{tabular}{cc cc cc}
\toprule
\multirow{2}{*}{\textbf{$d_{\text{ext}}$}} & \multirow{2}{*}{\shortstack{\textbf{Training} \\ \textbf{Fraction}}}
 & \multicolumn{2}{c}{\textbf{MS-SSIM $\uparrow$}} & \multicolumn{2}{c}{\textbf{LPIPS $\downarrow$}} \\
\cmidrule(lr){3-4} \cmidrule(lr){5-6}
 &  & \textbf{Type III} & \textbf{Type IV} & \textbf{Type III} & \textbf{Type IV} \\
\midrule
0.1 & 5\%   & 0.832 \sd{0.010} & 0.842 \sd{0.010} & 0.112 \sd{0.012} & 0.094 \sd{0.011} \\
0.1 & 40\%  & 0.834 \sd{0.010} & 0.844 \sd{0.010} & 0.110 \sd{0.012} & 0.093 \sd{0.011} \\
0.1 & 100\% & 0.835 \sd{0.010} & 0.845 \sd{0.010} & 0.110 \sd{0.012} & 0.093 \sd{0.011} \\
\midrule
0.5 & 5\%   & 0.667 \sd{0.014} & 0.680 \sd{0.014} & 0.330 \sd{0.015} & 0.312 \sd{0.014} \\
0.5 & 40\%  & 0.672 \sd{0.014} & 0.684 \sd{0.014} & 0.325 \sd{0.015} & 0.309 \sd{0.014} \\
0.5 & 100\% & 0.674 \sd{0.014} & 0.685 \sd{0.014} & 0.325 \sd{0.015} & 0.309 \sd{0.014} \\
\midrule
1.0 & 5\%   & 0.170 \sd{0.014} & 0.178 \sd{0.014} & 0.601 \sd{0.014} & 0.586 \sd{0.013} \\
1.0 & 40\%  & 0.172 \sd{0.014} & 0.179 \sd{0.014} & 0.599 \sd{0.014} & 0.585 \sd{0.013} \\
1.0 & 100\% & 0.173 \sd{0.014} & 0.180 \sd{0.014} & 0.599 \sd{0.014} & 0.585 \sd{0.013} \\
\bottomrule
\end{tabular}
\end{table}

The results of the fidelity–diversity analysis are provided in Fig.~\ref{fig:fidelity_diversity} and summarized in Tables~\ref{tab:crc_sim} and \ref{tab:pgi_sim}, which report MS-SSIM and LPIPS across the denoising extent $d_{\text{ext}}$ for both the RGB Colonoscopy dataset and the VTS GC Phantom dataset under three different training fractions (5\%, 40\%, 100\%). Several consistent trends emerge.

First, increasing $d_{\text{ext}}$ reliably decreases MS-SSIM while increasing LPIPS across all configurations. This monotonic behavior directly reflects the design of the guided sampling process: small $d_{\text{ext}}$ values correspond to starting from nearly clean latents, yielding high structural fidelity, whereas large $d_{\text{ext}}$ values correspond to starting from highly noisy latents, driving more diverse synthesis but substantially lower fidelity. For example, in the RGB Colonoscopy dataset (with 100\% training data, for the adenomatous class), MS-SSIM drops from $0.917 \pm 0.040$ at $d_{\text{ext}}=0.1$ to $0.159 \pm 0.125$ at $d_{\text{ext}}=1.0$, while LPIPS rises from $0.089 \pm 0.036$ to $0.655 \pm 0.059$ (Table~\ref{tab:crc_sim}). A similar trend is observed in the VTS GC Phantom dataset across the four Borrmann types, where MS-SSIM decreases from $0.822$--$0.845$ at $d_{\text{ext}}=0.1$ to $0.159$--$0.180$ at $d_{\text{ext}}=1.0$, accompanied by an increase in LPIPS from $0.093$--$0.123$ to $0.585$--$0.617$ (Table~\ref{tab:pgi_sim}).

Second, training fraction strongly modulates robustness, but the effect is dataset dependent. In the RGB Colonoscopy dataset, the similarity trends remain stable across training fractions: even with only 5\% of the data, the model achieves MS-SSIM and LPIPS values close to those observed with 40\% or 100\%. In contrast, the VTS GC Phantom dataset shows clear separation between fractions. At 5\%, MS-SSIM decays more steeply, particularly in the low-$d_{\text{ext}}$ regime where structural priors dominate. For instance, at $d_{\text{ext}}=0.1$, Borrmann~III cases yield MS-SSIM $0.832 \pm 0.010$ at 5\% versus $0.845 \pm 0.010$ at 100\% (Table~\ref{tab:pgi_sim}). With 40\% of the data, the trends largely stabilize and overlap those observed at 100\%, suggesting that the VTS GC Phantom dataset requires at least moderate training size to preserve reliable detail.

Third, in addition to the mean trends, the observed variance (Fig. \ref{fig:fidelity_diversity}) in MS-SSIM and LPIPS increases systematically with $d_{\text{ext}}$, particularly for MS-SSIM, reflecting that outputs become more diverse as sampling starts from noisier latents. At low $d_{\text{ext}}$, generations remain close to the input image and show little variability, while at high $d_{\text{ext}}$ the same input can yield substantially different reconstructions, resulting in broader variance bands. Notably, the RGB Colonoscopy dataset exhibits higher variance overall than the VTS GC Phantom dataset, which can be attributed to its larger scale and stronger models that capture more modes of variation. By contrast, the structurally simpler VTS GC Phantom dataset shows narrower variance at moderate and high training fractions, and marked instability at 5\% in the low-$d_{\text{ext}}$ regime.

Taken together, these findings highlight three important implications. First, $d_{\text{ext}}$ provides a tunable parameter for navigating the fidelity–diversity trade-off, enabling controlled transitions from faithful reconstructions to highly diverse generations. Second, the interaction between training fraction and dataset type is asymmetric: the RGB Colonoscopy dataset shows comparatively stable trends across training fractions, while the VTS GC Phantom dataset shows greater sensitivity to reduced training size, with more stable detail preservation emerging at moderate-to-higher training fractions. Third, the growth in variance with increasing $d_{\text{ext}}$ reflects enhanced diversity, with larger datasets leveraging this variance to explore richer generative modes and smaller datasets showing instability when data are scarce. These results collectively underscore the importance of dataset scale and class complexity in shaping both the mean and variability of generative outcomes.

\subsubsection{Role of $d_{\text{ext}}$ in Controlled Synthesis}

Taken together, the qualitative (Figs.~\ref{fig:crc-synth} and \ref{fig:pgi-synth}) and quantitative (Fig. \ref{fig:fidelity_diversity}, Tables \ref{tab:crc_sim} and \ref{tab:pgi_sim}) analyses highlight the central role of $d_{\text{ext}}$ as a controllable parameter in synthetic generation. At low values, $d_{\text{ext}}$ emphasizes structural fidelity by anchoring generations to the input, making it well suited for tasks that require anatomically faithful images, such as evaluating training classifiers on subtle tumor morphologies. At high values, $d_{\text{ext}}$ shifts the balance toward greater intra-class diversity, enabling augmentation strategies that enrich training distributions and improve generalization. Importantly, intermediate values (e.g., $d_{\text{ext}}=0.3$–0.5) offer a modest compromise, preserving fine-grained clinical details while introducing sufficient variability to avoid overfitting. Moreover, dataset scale appears to modulate the effectiveness of this trade-off: large and heterogeneous datasets sustain stable synthesis across the full $d_{\text{ext}}$ range, whereas smaller datasets benefit most from lower $d_{\text{ext}}$, where reliance on the image prior helps maintain realism. 

Notably, as shown in Fig.~\ref{fig:fidelity_diversity}, the CC-LDM trained on only 40\% of the data achieves performance comparable to the full-dataset model in the low-$d_{\text{ext}}$ regime, even though global metrics such as FID/KID remain higher (see Fig.~\ref{fig:data_req}). This finding underscores the efficiency of controlled guidance: by anchoring generation to the reference image at low $d_{\text{ext}}$, the model can preserve clinically relevant structures and textures with substantially reduced training data, while still maintaining sufficient variability for augmentation.
Thus $d_{\text{ext}}$ emerges as a practical tuning parameter for tailoring synthetic generation to specific downstream goals, from anatomy-faithful reconstruction to diversity-driven augmentation. This motivates our third experiment (see Sec. \ref{sec:task}), where we explicitly incorporate $d_{\text{ext}}$ into an uncertainty-guided augmentation strategy for downstream classification. For these experiments, we use the CC-LDMs trained on 100\% of the available data to isolate the effect of augmentation design.

\begin{table*}[!t]
\centering
\caption{RGB Colonoscopy downstream task results (mean $\pm$ std over 5 runs).}
\label{tab:rgb_results}
\small
\setlength{\tabcolsep}{4pt}
\begin{tabular}{l c cc cc cc}
\toprule
\multirow{2}{*}{\textbf{SDA Method}} & \multirow{2}{*}{$\boldsymbol{d_{\text{ext}}}$} & \multicolumn{2}{c}{\textbf{Dilated ResNet}} & \multicolumn{2}{c}{\textbf{ConvNeXt}} & \multicolumn{2}{c}{\textbf{Swin Transformer}} \\
\cmidrule(lr){3-4}\cmidrule(lr){5-6}\cmidrule(lr){7-8}
 &  & Acc. (\%) $\uparrow$ & F1 $\uparrow$ & Acc. (\%) $\uparrow$ & F1 $\uparrow$ & Acc. (\%) $\uparrow$ & F1 $\uparrow$ \\
\midrule
Real Data Only & -   & 62.8 \sd{1.6} & 0.670 \sd{0.014} & 67.4 \sd{3.6} & 0.725 \sd{0.030} & 67.9 \sd{2.9} & 0.731 \sd{0.029} \\
LC OS & -   & 63.1 \sd{1.1} & 0.675 \sd{0.012} & 68.9 \sd{1.8} & 0.738 \sd{0.018} & 66.1 \sd{5.7} & 0.711 \sd{0.058} \\
HE OS & -   & 64.5 \sd{2.5} & 0.684 \sd{0.023} & 68.7 \sd{3.7} & 0.735 \sd{0.033} & 65.0 \sd{4.6} & 0.705 \sd{0.042} \\
Label Cond.    & 1.0 & 64.2 \sd{1.8} & 0.677 \sd{0.012} & 75.8$^*$ \sd{1.6} & 0.796$^*$ \sd{0.018} & 74.9$^*$ \sd{2.3} & 0.786$^*$ \sd{0.021} \\
+ LC Guided    & 0.3 & 61.9 \sd{2.8} & 0.662 \sd{0.021} & 74.3$^*$ \sd{1.2} & 0.778$^*$ \sd{0.015} & 75.4$^*$ \sd{2.7} & \textbf{0.795$^*$ \sd{0.033}} \\
                & 0.5 & 62.5 \sd{3.2} & 0.665 \sd{0.024} & 72.2 \sd{1.2} & 0.763$^*$ \sd{0.005} & 73.6 \sd{2.9} & 0.778 \sd{0.028} \\
                & 0.7 & 64.0 \sd{4.1} & 0.678 \sd{0.035} & 75.0$^*$ \sd{2.9} & 0.792$^*$ \sd{0.026} & 73.8 \sd{3.4} & 0.778 \sd{0.034} \\
+ HE Guided    & 0.3 & 65.3$^*$ \sd{2.5} & 0.689$^*$ \sd{0.016} & 75.1$^*$ \sd{1.3} & 0.790$^*$ \sd{0.015} & 74.7$^*$ \sd{2.8} & 0.787$^*$ \sd{0.033} \\
                & 0.5 & \textbf{66.1$^*$ \sd{1.7}} & \textbf{0.694$^*$ \sd{0.013}} & 76.0$^*$ \sd{4.1} & 0.802$^*$ \sd{0.034} & \textbf{75.7$^*$ \sd{2.1}} & 0.787$^*$ \sd{0.021} \\
                & 0.7 & 64.6 \sd{2.4} & 0.687 \sd{0.022} & \textbf{77.2$^*$ \sd{1.2}} & \textbf{0.811$^*$ \sd{0.015}} & 74.4$^*$ \sd{2.3} & 0.782$^*$ \sd{0.019} \\
\bottomrule
\end{tabular}
\end{table*}

\begin{table*}[!t]
\centering
\caption{VTS GC Phantom downstream task results (mean $\pm$ std over 5 runs).}
\label{tab:vts_results}
\small
\setlength{\tabcolsep}{4pt}
\begin{tabular}{l c cc cc cc}
\toprule
\multirow{2}{*}{\textbf{SDA Method}} & \multirow{2}{*}{$\boldsymbol{d_{\text{ext}}}$} & \multicolumn{2}{c}{\textbf{Dilated ResNet}} & \multicolumn{2}{c}{\textbf{ConvNeXt}} & \multicolumn{2}{c}{\textbf{Swin Transformer}} \\
\cmidrule(lr){3-4}\cmidrule(lr){5-6}\cmidrule(lr){7-8}
 &  & Acc. (\%) $\uparrow$ & F1 $\uparrow$ & Acc. (\%) $\uparrow$ & F1 $\uparrow$ & Acc. (\%) $\uparrow$ & F1 $\uparrow$ \\
\midrule
Real Data Only & -   & 76.5 \sd{2.5} & 0.833 \sd{0.026} & 84.7 \sd{2.5} & 0.896 \sd{0.020} & 87.3 \sd{1.3} & 0.916 \sd{0.010} \\
LC OS & -   & 71.2 \sd{1.2} & 0.762 \sd{0.017} & 86.7 \sd{3.4} & 0.911 \sd{0.030} & 89.3 \sd{1.3} & \textbf{0.933$^*$ \sd{0.010}} \\
HE OS & -   & 71.9 \sd{1.5} & 0.771 \sd{0.022} & 84.7 \sd{3.8} & 0.894 \sd{0.036} & 87.3 \sd{1.1} & 0.919 \sd{0.010} \\
Label Cond.    & 1.0 & 83.8$^*$ \sd{1.7} & 0.902$^*$ \sd{0.012} & 86.3 \sd{2.1} & 0.907 \sd{0.014} & 88.4 \sd{1.7} & 0.924 \sd{0.015} \\
+ LC Guided    & 0.3 & 85.6$^*$ \sd{3.4} & 0.907$^*$ \sd{0.024} & 85.9 \sd{3.6} & 0.902 \sd{0.029} & 89.6 \sd{2.4} & \textbf{0.933$^*$ \sd{0.015}} \\
                & 0.5 & \textbf{85.6$^*$ \sd{2.4}} & \textbf{0.908$^*$ \sd{0.017}} & 86.2 \sd{2.9} & 0.903 \sd{0.022} & \textbf{89.7 \sd{2.1}} & 0.931 \sd{0.019} \\
                & 0.7 & 85.0$^*$ \sd{3.8} & 0.904$^*$ \sd{0.032} & 85.4 \sd{2.3} & 0.899 \sd{0.017} & 87.7 \sd{2.2} & 0.917 \sd{0.016} \\
+ HE Guided    & 0.3 & 84.2$^*$ \sd{3.3} & 0.896$^*$ \sd{0.025} & 84.9 \sd{2.4} & 0.895 \sd{0.022} & 88.7 \sd{2.3} & 0.927 \sd{0.015} \\
                & 0.5 & 84.0$^*$ \sd{2.7} & 0.897$^*$ \sd{0.019} & \textbf{88.1$^*$ \sd{0.5}} & \textbf{0.922$^*$ \sd{0.004}} & 88.1 \sd{2.4} & 0.925 \sd{0.015} \\
                & 0.7 & 85.3$^*$ \sd{3.4} & 0.906$^*$ \sd{0.026} & 86.4 \sd{2.3} & 0.907 \sd{0.020} & 88.7 \sd{2.2} & 0.926 \sd{0.014} \\
\bottomrule
\end{tabular}
\end{table*}

\subsection{Downstream Task Performance}

Results for the downstream classification task (see Sec. \ref{sec:task}) are summarized in Tables \ref{tab:rgb_results} and \ref{tab:vts_results}. Across both imaging modalities, depending on $d_{ext}$ and the uncertainty criterion, Uncertainty-Guided SDA yielded statistically significant improvements ($p < 0.05$) relative to the real-only baseline, while also outperforming label-only augmentation. On the other hand, simply replicating high-uncertainty samples (via HE/LC-based oversampling) often led to overfitting and degraded performance. This contrast highlights the benefits of generative synthesis over naive sample repetition to improve decision boundaries. 

When applying Uncertainty-Guided SDA, the Dilated ResNet, which struggled the most on both baseline datasets, benefited considerably from targeted augmentation. On the RGB colonoscopy dataset, HE selection at an intermediate denoising extent of $d_{ext}=0.5$ yielded a statistically significant improvement, boosting accuracy by 3.3\% (to 66.1\%) and F1 by 0.024 (to 0.694) relative to the baseline. ConvNeXt, a stronger and modern convolutional architecture, also demonstrated substantial gains, for e.g., leveraging HE guidance at a higher denoising extent ($d_{\text{ext}}=0.7$) yielded 77.2\% accuracy on the RGB data, while achieving 88.1\% accuracy on the VTS data at $d_{\text{ext}}=0.5$. For the Swin Transformer, gains from synthetic augmentation were more context-dependent. On the RGB colonoscopy dataset, it achieved a peak F1 of 0.795 with Low Confidence (LC) selection at $d_{\text{ext}}=0.3$. However, on the VTS GC phantom dataset, the model encountered a performance ceiling. The Swin baseline was already highly performant on this dataset (87.3\% accuracy, 0.916 F1), and guided augmentation (0.933 F1 at LC $d_{\text{ext}}=0.3$) only marginally matched the performance of the LC oversampling case.

The optimal configuration for Uncertainty-Guided SDA proved highly architecture-dependent. For the uncertainty metric, the Swin Transformer consistently achieved peak F1 scores using LC selection on both the RGB (0.795) and VTS (0.933) datasets, suggesting attention mechanisms benefit most from targeting under-confident instances. In contrast, ConvNeXt maximized both accuracy and F1 across both domains (RGB: 77.2\% Acc, 0.811 F1; VTS: 88.1\% Acc, 0.922 F1) using HE selection to resolve inherently ambiguous cases. The ideal $d_{\text{ext}}$ followed similar architectural constraints. Swin preferred lower generative diversity ($d_{\text{ext}}=0.3$), relying on high structural fidelity to the reference images. The Dilated ResNet required a balanced approach, peaking at an intermediate $d_{\text{ext}}=0.5$ for both datasets (0.694 RGB F1, 0.908 VTS F1). Conversely, the robust ConvNeXt successfully leveraged much higher diversity, peaking at $d_{\text{ext}}=0.7$ on the RGB data and $d_{\text{ext}}=0.5$ on the VTS data. 

Taken together, these results demonstrate that uncertainty-guided SDA offers a distinct advantage over purely label-guided SDA. While label-only augmentation risks generating biased data that reflects only generic prototypes, GUIDE helps mitigate this risk by anchoring novel variations to specific, challenging training examples. By targeting the regions of the feature space where a classifier is least reliable, the framework produces diverse, diagnostically relevant data. While performance plateaus naturally exist for already strong models, the substantial gains observed across architectures validate GUIDE as a practical pipeline for reducing dataset bias and improving generalization in GI tumor classification.

\subsection{Component Ablation}
Tables~\ref{tab:ablation_rgb_f1} and ~\ref{tab:ablation_vts_f1} summarize the results of the component ablation study (Sec.~\ref{sec:ablation}), showing that isolating individual components of the GUIDE SDA pipeline frequently yielded suboptimal performance. Applying sample selection in isolation (HE OS) consistently underperformed the real-only baseline across most configurations, confirming that resampling without generative variation is insufficient. Introducing unconstrained generative augmentation (Label-Only Conditioned Generation) immediately reversed this trend, providing a substantial performance lift and yielding statistically significant improvements for architectures like ConvNeXt (0.796 F1) and the Swin Transformer (0.786 F1) on the RGB dataset. However, integrating image guidance to complete the full GUIDE framework achieved the highest overall F1-scores in five of the six evaluated configurations. This combined approach drove peak, statistically significant performance for ConvNeXt across both domains (0.802 RGB, 0.922 VTS) and for the Swin Transformer on both datasets (0.787 RGB, 0.925 VTS). The sole exception was the Dilated ResNet on the VTS dataset, where unconstrained generation marginally outperformed the full pipeline (0.902 vs. 0.897 F1). Overall, the progression from baseline to full implementation confirmed that uniting targeted selection with image-guided synthesis generally produced the strongest performance.

\begin{table}[!t]
\centering
\caption{Component ablation on the RGB Colonoscopy dataset (F1-score $\uparrow$, mean $\pm$ std over 5 runs). }
\label{tab:ablation_rgb_f1}
\small
\setlength{\tabcolsep}{3.5pt}
\begin{tabular}{l c c c ccc}
\toprule
\multirow{3}{*}{\textbf{SDA Method}} & \multicolumn{3}{c}{\textbf{Components}} & \multirow{3}{*}{\textbf{\makecell{Dilated\\ResNet}}} & \multirow{3}{*}{\textbf{ConvNeXt}} & \multirow{3}{*}{\textbf{\makecell{Swin\\Trans.}}} \\
\cmidrule(lr){2-4}
& \makecell{Sample\\Sel.} & \makecell{Generative\\Aug.} & \makecell{Image\\Guid.} &  &  &  \\
\midrule
Real Data Only                  & $\times$     & $\times$     & $\times$     & 0.670 \sd{0.014} & 0.725 \sd{0.030} & 0.731 \sd{0.029} \\
HE OS                           & $\checkmark$ & $\times$     & $\times$     & 0.684 \sd{0.023} & 0.735 \sd{0.033} & 0.705 \sd{0.042} \\
Label-Conditioned Generation    & $\times$     & $\checkmark$ & $\times$     & 0.677 \sd{0.012} & 0.796$^*$ \sd{0.018} & 0.786$^*$ \sd{0.021} \\
GUIDE (HE, $d_{\text{ext}}=0.5$) & $\checkmark$ & $\checkmark$ & $\checkmark$ & \textbf{0.694$^*$ \sd{0.013}} & \textbf{0.802$^*$ \sd{0.034}} & \textbf{0.787$^*$ \sd{0.021}} \\
\bottomrule
\end{tabular}
\end{table}

\begin{table}[!t]
\centering
\caption{Component ablation on the VTS GC Phantom dataset (F1-score $\uparrow$, mean $\pm$ std over 5 runs). }
\label{tab:ablation_vts_f1}
\small
\setlength{\tabcolsep}{3.5pt}
\begin{tabular}{l c c c ccc}
\toprule
\multirow{3}{*}{\textbf{SDA Method}} & \multicolumn{3}{c}{\textbf{Components}} & \multirow{3}{*}{\textbf{\makecell{Dilated\\ResNet}}} & \multirow{3}{*}{\textbf{ConvNeXt}} & \multirow{3}{*}{\textbf{\makecell{Swin\\Trans.}}} \\
\cmidrule(lr){2-4}
& \makecell{Sample\\Sel.} & \makecell{Generative\\Aug.} & \makecell{Image\\Guid.} &  &  &  \\
\midrule
Baseline                        & $\times$     & $\times$     & $\times$     & 0.833 \sd{0.026} & 0.896 \sd{0.020} & 0.916 \sd{0.010} \\
HE OS                           & $\checkmark$ & $\times$     & $\times$     & 0.771 \sd{0.022} & 0.894 \sd{0.036} & 0.919 \sd{0.010} \\
Label-Conditioned Generation    & $\times$     & $\checkmark$ & $\times$     & \textbf{0.902$^*$ \sd{0.012}} & 0.907 \sd{0.014} & 0.924 \sd{0.015} \\
GUIDE (HE, $d_{\text{ext}}=0.5$) & $\checkmark$ & $\checkmark$ & $\checkmark$ & 0.897$^*$ \sd{0.023} & \textbf{0.922$^*$ \sd{0.004}} & \textbf{0.925 \sd{0.015}} \\
\bottomrule
\end{tabular}
\end{table}

\section{Conclusion}
In this work, we introduced and validated GUIDE, a weakly supervised diffusion-based augmentation framework for GI imaging that combines uncertainty-based sample selection with class- and image-conditioned generation. A key feature of GUIDE is its ability to control the fidelity–diversity trade-off through the tunable denoising extent, $d_{\text{ext}}$. We evaluated GUIDE across two datasets with very different imaging modalities (RGB images of colonoscopy videos and unique VTS images of GC tumors), highlighting the versatility of this approach. Low $d_{\text{ext}}$ yields anatomy-faithful images anchored to the reference, high $d_{\text{ext}}$ promotes intra-class diversity, and intermediate values often provide an effective balance. Notably, even with reduced training data, low $d_{\text{ext}}$ preserves clinically relevant structures, narrowing the gap to full-data models. We further demonstrated that uncertainty-guided synthetic data augmentation based on the targeted generation of ``hard'' training samples improves downstream classification across convolutional, transformer, and hybrid architectures, often outperforming class-only augmentation. These gains are generally strongest when synthetic images remain close to the real data distribution, corresponding to low-to-mid $d_{\text{ext}}$ settings, although stronger backbones can in some cases benefit from the additional diversity introduced at higher $d_{\text{ext}}$. Overall, GUIDE provides a controllable and data-efficient strategy for generating diagnostically relevant synthetic images in limited-data GI imaging scenarios. Since GUIDE operates with only class labels, it is readily applicable in settings where dense annotations are unavailable. Future work could extend it to flexibly exploit varying levels of annotation (e.g., boxes, masks, metadata) when available, enabling finer control without losing scalability.  Furthermore, alternative more sophisticated strategies for selecting samples to augment training, including margin sampling \cite{scheffer2001active}, 
influence- or gradient-based scoring (e.g., TracIn \cite{pruthi2020estimating}), and data valuation approaches such as Shapley values \cite{ghorbani2019data,jia2019towards}, can also be explored.

\section*{Disclosures}

Farshid Alambeigi received financial support from the National Institutes of Health (NCI), the Andrew Sabin Family Foundation, Project Perwinkle, The IC2 Institute (UT Austin), and Texas Robotics Seed Collaborative Funding. Additionally, he has a pending patent for Reliability Assessment Analysis and Calibration for Artificial Intelligence Classification with the Board of Regents, The University of Texas System. All other authors declare no competing financial interests or personal relationships that could have influenced this work.

During the preparation of this work the author(s) used ChatGPT (OpenAI) and  Gemini (Google) in order to assist with drafting, language refinement, and literature organization. After using these tools, the author(s) reviewed and edited the content as needed and take(s) full responsibility for the content of the published article.

\section*{Code, Data, and Materials Availability}
This paper utilizes two datasets to support its findings. The dataset of RGB colonoscopy images of CRC tumors is publicly available via the Harvard Dataverse (DOI: 10.7910/DVN/FCBUOR). The unique VTS dataset of GC phantoms was collected in-house using the HySenSe sensor and is not publicly available. Details regarding the data collection procedure and phantom fabrication are provided in the manuscript and in referenced prior works. While the code utilized for the GUIDE framework is not publicly available, the latent diffusion process, sequence-aware conditioning, and uncertainty-guided sampling strategy, as well as key hyperparameters, are described in detail within the manuscript to facilitate the reproduction of the reported results.

\section*{Acknowledgements}
This work was supported by the National Cancer Institute of the National Institutes of Health under Award Number R21CA280747, Andrew Sabin Family Foundation and Project Periwinkle, the IC$^{2}$ Institute, and Texas Robotics Seed Collaborative Funding at the University of Texas at Austin.


\bibliography{guide}   
\bibliographystyle{spiejour}   




\vspace{1ex}
\noindent Biographies and photographs of the other authors are not available.

\listoffigures
\listoftables

\end{spacing}
\end{document}